\documentclass{article}

    \PassOptionsToPackage{numbers, compress}{natbib}

\usepackage[preprint]{neurips_2019}

\usepackage[utf8]{inputenc} 
\usepackage[T1]{fontenc}    
\usepackage[hidelinks]{hyperref}       
\usepackage{url}            
\usepackage{booktabs}       
\usepackage{tabularx}
\usepackage{amsfonts}       
\usepackage{nicefrac}       
\usepackage{microtype}      
\usepackage{bm}
\usepackage{amsmath}
\usepackage{color} 
\usepackage{algorithm,algorithmic}

\usepackage{wrapfig}
\usepackage{graphicx}
\usepackage{caption}
\usepackage{subcaption}
\graphicspath{{./images/}}

\usepackage[disable]{todonotes}
\newcommand{\acN}[1]{\todo[size=\scriptsize, color=blue!30]{[AC] #1}}

\newcommand{\srN}[1]{\todo[size=\scriptsize, color=cyan]{[SR] #1}}

\usepackage[many]{tcolorbox} 
\definecolor{mycolor}{rgb}{0.122, 0.435, 0.698}
\newtcolorbox{mybox}[1]{breakable, colback=mycolor!5!white, colframe=white}

\title{Introducing an Explicit Symplectic Integration Scheme for Riemannian Manifold Hamiltonian Monte Carlo}

\author{%
  Adam D. Cobb \\
  Department of Engineering Science\\
  University of Oxford\\
  \texttt{acobb@robots.ox.ac.uk} \\
  \And
  Atılım Güneş Baydin\\
  Department of Engineering Science\\
  University of Oxford\\
  \And
  Andrew Markham \\
  Department of Computer Science \\
  University of Oxford \\
  \And
  Stephen J. Roberts \\
  Department of Engineering Science\\
  University of Oxford\\
}

\begin{document}

\maketitle

\begin{abstract}
We introduce a recent symplectic integration scheme derived for solving physically motivated systems with non-separable Hamiltonians. We show its relevance to Riemannian manifold Hamiltonian Monte Carlo (RMHMC) and provide an alternative to the currently used generalised leapfrog symplectic integrator,
which relies on solving multiple fixed point iterations to convergence. Via this approach, we are able to reduce the number of higher-order derivative calculations per leapfrog step. We explore the implications of this integrator and demonstrate its efficacy in reducing the computational burden of RMHMC. Our code is provided in a new open-source Python package, \texttt{hamiltorch}. \acN{New abstract edit!}\srN{have edited}
\end{abstract}

\section{Introduction}
In Bayesian statistics, we often find ourselves with the challenge of performing an integration over a set of model parameters $\bm{\omega}$. A common example is when we have a model, defined by the likelihood $p(\mathbf{y}\vert\mathbf{x},\bm{\omega})$, with $\mathbf{x}$ and $\mathbf{y}$ constituting an input--output data pair. We define a reasonable prior over the model parameters, $p(\bm{\omega})$ and our interest lies in inferring the posterior over the parameters $p(\bm{\omega}\vert\mathbf{x},\mathbf{y})$. Knowledge of this posterior allows us to make predictions over new input points $\mathbf{x}^*$ via the integration,
\begin{equation}\label{eq:pred}
 p(\mathbf{y}^*\vert\mathbf{x}^*,\mathbf{x},\mathbf{y}) = \int p(\mathbf{y}^*\vert\mathbf{x}^*,\bm{\omega})\,p(\bm{\omega}\vert\mathbf{x},\mathbf{y})\mathrm{d}\bm{\omega}.
\end{equation}
In Bayesian modelling, the path to this predictive distribution comes from employing Bayes' theorem:
\begin{equation}\label{eq:Bayes}
p(\bm{\omega}\vert\mathbf{x},\mathbf{y}) = \frac{p(\mathbf{y}\vert\mathbf{x},\bm{\omega})\,p(\bm{\omega})}{p(\mathbf{y}\vert\mathbf{x})},
\end{equation}
where the marginal likelihood, $p(\mathbf{y}\vert\mathbf{x})$, is the final term that needs to be inferred in order to obtain the predictive distribution in Equation \eqref{eq:pred}. This marginal likelihood also requires integration with respect to $\bm{\omega}$:
\begin{equation}\label{eq:ml}
p(\mathbf{y}\vert\mathbf{x}) = \int p(\mathbf{y}\vert\mathbf{x},\bm{\omega})\,p(\bm{\omega})\mathrm{d}\bm{\omega}.
\end{equation}
Therefore, if we can infer $p(\mathbf{y}\vert\mathbf{x})$, we can evaluate the predictive distribution. Depending on the prior and the likelihood, there are sometimes analytic solutions to marginalising over $\bm{\omega}$, such as in Gaussian process regression \citep{williams2006gaussian}, where the prior is conjugate to the likelihood. However, when this is not the case, one must approximate the integral.
One solution is to replace the true posterior with a variational approximation that is either cheaper to sample from (stochastic variational inference) \citep{hoffman2013stochastic} or conjugate to the likelihood \citep{mackay1992practical}. Another option is to use a Markov chain Monte Carlo (MCMC) technique to sample directly from the posterior in Equation \eqref{eq:Bayes}.

In this paper we focus on the latter option of using an MCMC technique, namely, Riemannian Manifold Hamiltonian Monte Carlo (RMHMC) \citep{girolami2011riemann}. As we describe in Sections \ref{sec:hmc} and \ref{sec:rmhmc}, RMHMC is a Monte Carlo technique that is well suited to performing inference in highly complex models.

However, RMHMC implementations remain computationally expensive and challenging due to both the higher-order derivatives and quadratic memory requirements. It is for these reasons that RMHMC has not become widely used in large-scale Bayesian modelling, such as in Bayesian neural networks \citep{mackay1992practical, neal2011mcmc}. To overcome these issues, we present a new RMHMC integration scheme, \textit{Explicit RMHMC}, that is competitive with standard Hamiltonian Monte Carlo (HMC) in both wall-clock time and performance. 
Furthermore, we provide a Python package, \texttt{hamiltorch}, that implements a general-purpose API to run these algorithms on a GPU with PyTorch \cite{paszke2017automatic}.\footnote{Link to the Python package: \url{https://github.com/AdamCobb/hamiltorch}. For a tutorial please refer to \url{https://adamcobb.github.io/journal/hamiltorch.html}.}

Our paper is structured as follows. Section \ref{sec:lit_review} reviews related literature, providing context for our work. Sections \ref{sec:hmc} and \ref{sec:rmhmc} provide theory, which are then followed by experimental results in Section~\ref{sec:exp}. Finally, we provide closing remarks in Section~\ref{sec:conclusion}. 


\section{Related work}\label{sec:lit_review}

Our motivation for introducing a more computationally efficient way of performing inference with RMHMC stems from quotes in the literature, such as \textit{``we need improved inference techniques in BNNs"}, \citet{gal2018sufficient}; and \textit{``In small scale problems, Markov chain Monte Carlo algorithms remain the gold standard, but such algorithms face major problems in scaling up to big data.''}, \citet{li2017simple}. Both these papers are referring to the need for improved inference in complex models. Furthermore they also highlight the drawbacks of variational approximations \cite{blei2017variational}, which can lead to poor uncertainty quantification in safety-critical applications \citep{cobb2018loss} and possibly be more susceptible to adversarial attacks \citep{gal2018sufficient}.

\subsection{Markov chain Monte Carlo methods}

One of the most commonly used and most popularised MCMC algorithms originated from the statistical mechanics literature: the Metropolis--Hastings (MH) algorithm \citep{metropolis1953equation,hastings1970monte} includes an acceptance step, whereby proposed samples are accepted according to an acceptance probability that ensures the required MCMC properties are satisfied.\footnote{This acceptance step was originally related to potential energy of particles in a system, such that a sample configuration of particles was accepted according to this potential energy, rather than the previous method of simply weighting samples by the energy distribution.} Despite its success, the MH algorithm can suffer in high dimensional models due to low acceptance rates and an inability to efficiently explore the space (\citet[Chapter~11]{Bishop2006}).

HMC has become a popular choice for performing inference in highly complex models, due to its higher acceptance rates and its use of gradient information in exploring the space. It was first introduced as ``Hybrid Monte Carlo'' by \citet{duane1987hybrid} for computer simulations in lattice field theory.\footnote{We note that earlier work by \citet{alder1959studies} utilised Hamiltonian dynamics to simulate molecular dynamics.} However the term ``Hamiltonian Monte Carlo'' was coined from the extensive work by \citet{neal1995bayesian}, who introduced HMC to the machine learning community with work on Bayesian neural networks. Despite the success of HMC, there are various hyperparameters that need to be tuned, such as the mass matrix, $\mathbf{M}$, the step size, $\epsilon$, and the number of steps in a given trajectory, $L$ (or leapfrog steps, see Section \ref{sec:hmc}). The reliance on heuristics \cite{neal1995bayesian}, which are heavily dependent on the Bayesian model or the data, often prevent HMC from being exploited to its full potential.

Progress in this area has been made with the introduction of a Hamiltonian-based Monte Carlo scheme that takes into account the geometry of the Bayesian model. This was framed in the work by \citet{girolami2011riemann}, where they introduced RMHMC as an \textit{``overarching geometric framework''} for MCMC methods that included previous work by \citet{roberts2002langevin} and \citet{duane1987hybrid}.

\subsection{Adaptations to Hamiltonian-based Monte Carlo samplers}

The challenges associated with setting the hyperparameters of Hamiltonian samplers have seen various solutions offered in the literature. \citet{hoffman2014no} focussed on solving the problem of automatically setting the trajectory length $L$ in HMC by introducing the No-U-Turn Sampler, whilst ensuring detailed balance is preserved. This was extended to Riemannian manifolds in \citet{betancourt2013generalizing} with the aim of reducing unnecessary computation time. \citet{wang2013adaptive} take a different approach to adapting hyperparameters by employing Bayesian optimisation \citep{snoek2012practical} for adaptive tuning within Hamiltonian-based Monte Carlo samplers.

Rather than focussing on techniques for hyperparameter tuning, others have directly worked on the formulation of the Hamiltonian itself. \citet{shahbaba2014split} show that it is possible to split the Hamiltonian in a way that can reduce the computational cost. They use a MAP approximation that allows the Hamiltonian to be partially analytically integrated, which can reduce the computational burden when exploring much of the state space. 
Previous work by \citet{lan2012lagrangian} introduced some of the benefits of explicit integration techniques to RMHMC. However, unlike the integrator introduced in our work, they derive an integrator that is no longer symplectic and adjust their acceptance step in order to preserve detailed balance. In addition, they introduce two additional matrix inversions that limit efficiency in high-dimensional spaces.


\section{Hamiltonian Monte Carlo}\label{sec:hmc}

Rather than relying on a random walk to explore the parameter space of a model, HMC employs Hamiltonian dynamics to traverse the parameter space, $\bm{\omega}\in \mathbb{R}^D$. In order to introduce HMC, we begin with the integrand of interest $f(\bm{\omega}) = p(\mathbf{y} \vert \mathbf{x}, \bm{\omega})\,p(\bm{\omega})$ and augment the space by introducing the momentum variable $\mathbf{p} \in \mathbb{R}^D$, such that we now have a joint density $f(\bm{\omega})\,p(\mathbf{p}) =  f(\bm{\omega})\,\mathcal{N}(\mathbf{p} \vert \mathbf{0}, \mathbf{M})$. The negative log joint probability is then given by
\begin{equation}\label{eq:ham}
H(\bm{\omega},\mathbf{p}) = -\log\left(f(\bm{\omega})\right) + \frac{1}{2}\log\left((2\pi)^D \vert \mathbf{M} \vert\right) + \frac{1}{2}\mathbf{p}^{\top}\mathbf{M}^{-1}\mathbf{p}.
\end{equation}
It is then noted that this negative joint likelihood actually consists of a kinetic energy term $\frac{1}{2}\mathbf{p}^{\top}\mathbf{M}^{-1}\mathbf{p}$ and a potential energy term $-\mathcal{L}(\bm{\omega}) = -\log\left(f(\bm{\omega})\right)$. This insight points to using Hamiltonian dynamics to explore this system, where trajectories in this space move along constant energy levels.

Moving in this space requires derivative information:
\begin{equation}\label{eq:HMC_deriv}
\frac{\mathrm{d}\bm{\omega}}{\mathrm{d}\tau} = \frac{\partial H}{\partial \mathbf{p}} = \mathbf{M}^{-1}\mathbf{p}; \quad \frac{\mathrm{d}\bm{p}}{\mathrm{d}\tau} = - \frac{\partial H}{\partial \bm{\omega}} = \nabla_{\bm{\omega}}\mathcal{L}(\bm{\omega}).
\end{equation}
The evolution of this system must preserve both volume and total energy, where any transformation that preserves area has the property of \textit{symplecticity} \citep[Page~182]{hairer2006geometric}. As this Hamiltonian is separable, meaning $\frac{\partial H}{\partial \mathbf{p}}$ and $\frac{\partial H}{\partial \bm{\omega}}$ are functions of only one of the variables (momentum and parameters, respectively), to traverse the space we can use the Stormer--Verlet or leapfrog integrator (as used in \citet{duane1987hybrid} and \citet{neal1995bayesian}):
\begin{equation}\label{eq:leapfrog}
\mathbf{p}_{\tau + \epsilon / 2}  = \mathbf{p}_{\tau} + \frac{\epsilon}{2}\frac{\mathrm{d}\bm{p}}{\mathrm{d}\tau}(\bm{\omega}_{\tau}),\quad 
\bm{\omega}_{\tau + \epsilon}  = \bm{\omega}_{\tau} + \epsilon \frac{\mathrm{d}\bm{\omega}}{\mathrm{d}\tau}(\mathbf{p}_{\tau + \epsilon / 2}),\quad 
\mathbf{p}_{\tau + \epsilon}  = \mathbf{p}_{\tau + \epsilon / 2} + \frac{\epsilon}{2}\frac{\mathrm{d}\bm{p}}{\mathrm{d}\tau}(\bm{\omega}_{\tau + \epsilon})\;,
\end{equation}
where $\tau$ corresponds to a the leapfrog step iteration and $\epsilon$ is the step size. Therefore given this is a numerical solution that uses discrete steps, we must check whether the Hamiltonian energy is conserved by employing a Metropolis--Hastings step with an acceptance probability $\mathrm{min} \{0, H_{\tau = 0} - H_{\tau = L}\}$ after $L$ leapfrog steps. Finally, the overall sampling process follows the Gibbs sampling structure, where we draw $\mathbf{p} \sim p(\mathbf{p})$ and then draw a new set of parameters $\bm{\omega}^* \sim p(\bm{\omega}^*\vert\mathbf{p}) \propto f(\bm{\omega}^*)\,p(\mathbf{p} + \delta\mathbf{p})$, where we have followed similar nomenclature to \citet{girolami2011riemann}.

Therefore in repeating this Gibbs sampling scheme, including the implementation of the leapfrog algorithm, we can sample our parameter of interest $\bm{\omega}$.

\section{Moving to a Riemannian manifold}\label{sec:rmhmc}

One of the big challenges, when implementing HMC, is how to set the mass matrix~$\mathbf{M}$. In \citet[Appendix~4]{neal1995bayesian}, one practical solution is offered, which relates the step size for each parameter to the second-order derivative of the log posterior. Interestingly, this is not too far away from the topic of discussion in this section, where we start to explore ideas of higher-order derivatives.

If our interest lies in measuring distances between distributions, then naively treating the space as if it were Euclidean (while computationally convenient) may prevent rapid sampling of the posterior. A common metric for measuring a distance between distributions is the Kullback--Leibler (KL) divergence:
\begin{equation}
\mathrm{D}_{\mathrm{KL}}\big(p(\mathbf{x}, \bm{\omega}) \vert\vert p(\mathbf{x}, \bm{\omega}')\big) = \iint p(\mathbf{x}, \bm{\omega}) \left(\log p(\mathbf{x}, \bm{\omega}) - \log p(\mathbf{x}, \bm{\omega}')\right) \mathrm{d}\mathbf{x}\mathrm{d}\bm{\omega}.
\end{equation}
If we expand the $\log p(\mathbf{x}, \bm{\omega}')$ term within the integrand by using a Taylor expansion around $\bm{\omega}$ such that $\delta\bm{\omega} = \bm{\omega}' - \bm{\omega}$, we get (see Appendix \ref{ap:metric} for full derivation)
\begin{align}\label{eq:metric}
\mathrm{D}_{\mathrm{KL}}\big(p(\mathbf{x}, \bm{\omega}) \vert\vert p(\mathbf{x},  \bm{\omega} + \delta\bm{\omega})\big) \approx & - \iint p(\mathbf{x}, \bm{\omega}) \left(\frac{1}{2} \delta\bm{\omega}^{\top} \nabla_{\bm{\omega}}^2 \log p(\mathbf{x}, \bm{\omega})\delta\bm{\omega}\right)\mathrm{d}\mathbf{x}\mathrm{d}\bm{\omega} \notag\\
= & - \iint p(\mathbf{x}, \bm{\omega}) \left(\frac{1}{2} \delta\bm{\omega}^{\top} \mathbf{G}(\mathbf{x}, \bm{\omega})\delta\bm{\omega}\right)\mathrm{d}\mathbf{x}\mathrm{d}\bm{\omega}\;,
\end{align}
where $\mathbf{G}(\mathbf{x}, \bm{\omega})$ tells us about the curvature for the joint probability space over $\mathbf{x}$ and $\bm{\omega}$. As used in \citet{girolami2011riemann}, we are using a Bayesian perspective and concern ourselves with the joint likelihood of the data and the parameters and therefore $\mathbf{G}(\bm{\omega})$ is the Fisher information plus the negative Hessian of the log-prior. Furthermore there is a plethora of literature relating to the appropriate metric for measuring distances on a Riemannian manifold, where we refer to either the RMHMC paper \citep{girolami2011riemann} or the book \citep{amari2000methods}.

\subsection{Riemannian manifold Hamiltonian Monte Carlo}
Having introduced the need for a metric situated on a Riemannian manifold, we can now go back to the original Hamiltonian of Equation \eqref{eq:ham} but write it so that it lies on a Riemannian manifold by replacing $\mathbf{M}$ with $\mathbf{G}(\bm{\omega})$. However, because the mass matrix is now parameterised by $\bm{\omega}$, the Hamiltonian equations are no longer separable as they contain a coupling between $\bm{\omega}$ and $\mathbf{p}$:
\begin{equation}
\frac{\mathrm{d}\bm{\omega}}{\mathrm{d}\tau} = \frac{\partial H}{\partial \mathbf{p}} = \mathbf{G}(\bm{\omega})^{-1}\mathbf{p} \quad \text{and} \quad \frac{\mathrm{d}\bm{p}}{\mathrm{d}\tau} = - \frac{\partial H}{\partial \bm{\omega}} = \nabla_{\bm{\omega}}\mathcal{L}(\bm{\omega}) - \nabla_{\bm{\omega}}\mathcal{N}(\mathbf{0},\mathbf{G}(\bm{\omega})^{-1})\;.
\end{equation}

Since the Hamiltonian is non-separable, the leapfrog integrator defined in Equation~\ref{eq:leapfrog} is no longer applicable due to the volume conservation requirements. 
Furthermore, the new dependence of the mass matrix on the parameters means that detailed balance would no longer be satisfied.\footnote{At this point, it feels important to highlight that in general, implementations of HMC are often not time-reversible, since that would require an equal chance of $\epsilon$ being positive or negative. However, as \citet{hoffman2014no} mention, this can be omitted in practice.}


\section{Explicit RMHMC}

\subsection{Implicit versus explicit integration}\label{sec:implicit}

The current solution for solving the non-separable Hamiltonian equations in Section~\ref{sec:rmhmc} is to rely on the \textit{implicit} generalised leapfrog algorithm \citep{leimkuhler2004simulating}. The term \textit{implicit} refers to the computationally expensive first-order implicit Euler integrators, which are fixed-point iterations that are run until convergence. In order to understand the computational cost associated with this generalised leapfrog algorithm, we summarise a single leapfrog step in Algorithm~\ref{alg:gen_leap}.

A single step in the implicit generalised leapfrog contains two `while' loops, which both require expensive update steps for each iteration. In \citet{betancourt2013general}, the author uses `while' loops as in Algorithm~\ref{alg:gen_leap} and sets a threshold $\delta$ for breaking out. As an alternative to a conditional break, \citet{girolami2011riemann} use a `for' loop and fix the number of iterations to be five or six. Therefore, one step, (at a minimum) requires $11$ Hessian calculations along with the additional derivative calculations. 

Rather than relying on implicit integration, an \textit{explicit} integration scheme relies on an explicit evaluation at an already known value \citep[Chapter~1, Page~3]{hairer2006geometric}. This leads to a deterministic number of operations per leapfrog step, which we will show to be advantageous. We now introduce an explicit integration scheme for RMHMC.

\subsection{Introducing the new explicit integrator}

The challenge in speeding up the symplectic integration for non-separable Hamiltonians has not been applied within the context of Bayesian statistics. However if we delve into the physical sciences, there have been many examples where non-separable Hamiltonians appear \cite{tao2016explicitElectro,luo2017explicit,zhang2018explicit}. Therefore it makes sense to draw from the progress made in this area, given the strong relationship between the physical interpretations of HMC within probabilistic modelling applications.

A key advance that moved away from dealing with specific subclasses of non-separable Hamiltonians and moved towards arbitrary non-separable Hamiltonians came in \citet{pihajoki2015explicit}, where the phase space was extended to include an additional momentum term $\tilde{\mathbf{p}}$ and parameter term $\tilde{\bm{\omega}}$. This extension of the phase space was built on by \citet{tao2016explicit} who added an additional binding term that resulted in improved long-term performance. This new augmented Hamiltonian can now be defined as:
\begin{equation}\label{eq:ham_tao}
\tilde{H}(\bm{\omega},\mathbf{p},\tilde{\bm{\omega}},\tilde{\mathbf{p}}) = H_1(\bm{\omega},\tilde{\mathbf{p}}) + H_2(\tilde{\bm{\omega}},\mathbf{p}) + \varOmega h(\bm{\omega},\mathbf{p},\tilde{\bm{\omega}},\tilde{\mathbf{p}})\;,
\end{equation}
where the binding term $h(\bm{\omega},\mathbf{p},\tilde{\bm{\omega}},\tilde{\mathbf{p}}) = \Vert \bm{\omega} - \tilde{\bm{\omega}} \Vert_2^2 / 2 + \Vert \mathbf{p} - \tilde{\mathbf{p}} \Vert_2^2 / 2$ and $\varOmega$ controls the binding. The two Hamiltonian systems $H_1$ and $H_2$ each follow Equation~\eqref{eq:ham}, where $H(\bm{\omega},\tilde{\mathbf{p}})$ and $H(\tilde{\bm{\omega}},\mathbf{p})$ have their parameters and momenta mixed. Interestingly, if we set $\varOmega = 0$, we retrieve the augmented Hamiltonian from \citet{pihajoki2015explicit}.

The Hamiltonian in Equation~\eqref{eq:ham_tao} gives the system:
\begin{equation}
\begin{array}{cc}
\dot{\bm{\omega}} =& \delta \partial_{\mathbf{p}} H(\tilde{\bm{\omega}},\mathbf{p}) + \varOmega (\mathbf{p}-\tilde{\mathbf{p}})\\
\dot{\mathbf{p}} =& - \delta \partial_{\bm{\omega}} H(\bm{\omega},\tilde{\mathbf{p}}) - \varOmega (\bm{\omega}-\tilde{\bm{\omega}})\\
\dot{\tilde{\bm{\omega}}} =& \delta \partial_{\tilde{\mathbf{p}}} H(\bm{\omega},\tilde{\mathbf{p}}) + \varOmega (\tilde{\mathbf{p}} - \mathbf{p})\\
\dot{\tilde{\mathbf{p}}}=& - \delta \partial_{\tilde{\bm{\omega}}} H(\tilde{\bm{\omega}},\mathbf{p}) - \varOmega (\tilde{\bm{\omega}} - \bm{\omega}).
\end{array}
\end{equation}

We now introduce the integrators from \citet{tao2016explicit}, which when combined make up the symplectic flow for the augmented non-separable Hamiltonian:
\begin{equation}
\phi^{\delta}_{H_1} :
\begin{bmatrix}
    \bm{\omega}\\
    \mathbf{p}\\
    \tilde{\bm{\omega}}\\
    \tilde{\mathbf{p}}
\end{bmatrix}
\mapsto
\begin{bmatrix}
    \bm{\omega}\\
    \mathbf{p} - \delta \partial_{\bm{\omega}} H(\bm{\omega},\tilde{\mathbf{p}})\\
    \tilde{\bm{\omega}} + \delta \partial_{\tilde{\mathbf{p}}} H(\bm{\omega}, \tilde{\mathbf{p}})\\
    \tilde{\mathbf{p}}
\end{bmatrix}, 
\quad
\phi^{\delta}_{H_2} :
\begin{bmatrix}
    \bm{\omega}\\
    \mathbf{p}\\
    \tilde{\bm{\omega}}\\
    \tilde{\mathbf{p}}
\end{bmatrix}
\mapsto
\begin{bmatrix}
    \bm{\omega} + \delta \partial_{\mathbf{p}} H(\tilde{\bm{\omega}},\mathbf{p})\\
    \mathbf{p}\\
    \tilde{\bm{\omega}}\\
    \tilde{\mathbf{p}} - \delta \partial_{\tilde{\bm{\omega}}} H(\tilde{\bm{\omega}},\mathbf{p})\\
\end{bmatrix},\notag
\end{equation}
\begin{equation}\label{eq:tao_linear_eq}
\phi^{\delta}_{\varOmega h} :
\begin{bmatrix}
    \bm{\omega}\\
    \mathbf{p}\\
    \tilde{\bm{\omega}}\\
    \tilde{\mathbf{p}}
\end{bmatrix}
\mapsto
\frac{1}{2}
\begin{bmatrix}
	\begin{pmatrix}
	\bm{\omega}+\tilde{\bm{\omega}}\\
    \mathbf{p}+\tilde{\mathbf{p}}
	\end{pmatrix}
	+ \mathbf{R}(\delta)	
	\begin{pmatrix}
	\bm{\omega}-\tilde{\bm{\omega}}\\
    \mathbf{p}-\tilde{\mathbf{p}}
	\end{pmatrix}
	\\
    \begin{pmatrix}
	\bm{\omega}+\tilde{\bm{\omega}}\\
    \mathbf{p}+\tilde{\mathbf{p}}
	\end{pmatrix}
	- \mathbf{R}(\delta)	
	\begin{pmatrix}
	\bm{\omega}-\tilde{\bm{\omega}}\\
    \mathbf{p}-\tilde{\mathbf{p}}
	\end{pmatrix}
\end{bmatrix},
\quad
\text{where }
\mathbf{R}(\delta) = 
\begin{bmatrix}
    \cos (2\varOmega\delta)\mathbf{I} & \sin (2\varOmega\delta)\mathbf{I}\\
    -\sin (2\varOmega\delta)\mathbf{I} & \cos (2\varOmega\delta)\mathbf{I}
\end{bmatrix}.
\end{equation}
Finally, the numerical symplectic second-order integrator is given by the function composition
\begin{equation}\label{eq:tao_integ}
\phi^{\delta}_{\tilde{H}} = \phi^{\delta/2}_{H_1} \circ  \phi^{\delta/2}_{H_2} \circ \phi^{\delta}_{\varOmega h} \circ \phi^{\delta/2}_{H_2} \circ  \phi^{\delta/2}_{H_1},
\end{equation}
where combining these symmetric mappings is known as Strang splitting \citep{strang1968construction} and higher-order integrators can be built in a similar manner \citep{yoshida1990construction}. In Appendix \ref{ap:check}, we show that this second-order integrator is indeed symplectic.

There are two major advantages when comparing the integrator in Equation~\eqref{eq:tao_integ} to previous symplectic integrators for non-separable Hamiltonians. First of all, when comparing to the implicit integrator of Section \ref{sec:implicit}, we no longer need to rely on two fixed-point methods that can often diverge. Furthermore we are guaranteed the same number of derivative calculations of eight per equivalent leapfrog step. This compares to the minimum number of $11$ for the implicit generalised leapfrog (although it can often be higher).
Secondly, when comparing to the previous explicit integrator of \citet{pihajoki2015explicit}, this integration scheme has better long-term error performance. More specifically, \citet{tao2016explicit} showed that until the number of steps, $N = \mathcal{O}(\mathrm{min}(\delta^{-2}\varOmega^{-1}, \varOmega^{^1/_2}))$, the numerical error is  $\mathcal{O}(N\delta^{2}\varOmega)$, although empirical results show that larger values of $N$ give favourable results in terms of the sampler's performance.

\begin{minipage}[t]{0.49\textwidth}
  \begin{algorithm}[H]
    \centering
    \caption{Implicit Leapfrog Step}
    \label{alg:gen_leap}
    \begin{algorithmic}[1]
      \scriptsize
      \STATE \textbf{Inputs:} $ \mathbf{p}_0  $, $ \bm{\omega}_0, \epsilon, $
      \STATE $\mathbf{p} = \mathbf{p}_0 $
      \WHILE{$\Delta p > \delta $}
      \STATE $\mathbf{p}' = \mathbf{p}_0 + \frac{\epsilon}{2}\frac{\mathrm{d}\bm{p}}{\mathrm{d}\tau}(\mathbf{p},\bm{\omega}_{0})$
      \STATE $\Delta p = \mathrm{max}_i \{ \vert p_i - p_i'\vert\}$
      \STATE $\mathbf{p} = \mathbf{p}' $
      \ENDWHILE
      \STATE $\bm{\omega} = \bm{\omega}_0 $
      \WHILE{$\Delta \omega > \delta $}
      \STATE $\bm{\omega}' = \bm{\omega}_0 + \frac{\epsilon}{2}\frac{\mathrm{d}\bm{\omega}}{\mathrm{d}\tau}(\mathbf{p},\bm{\omega}_{0}) + \frac{\epsilon}{2}\frac{\mathrm{d}\bm{\omega}}{\mathrm{d}\tau}(\bm{p},\bm{\omega})$
      \STATE $\Delta \omega = \mathrm{max}_i \{ \vert \omega_i - \omega_i'\vert\}$
      \STATE $\bm{\omega} = \bm{\omega}' $
      \ENDWHILE
      \STATE $\mathbf{p} = \mathbf{p} + \frac{\epsilon}{2}\frac{\mathrm{d}\bm{p}}{\mathrm{d}\tau}(\mathbf{p},\bm{\omega})$\\\vspace{1.5mm}
    \end{algorithmic}
  \end{algorithm}
\end{minipage}\hfill
\begin{minipage}[t]{0.49\textwidth}
   \begin{algorithm}[H]
      \centering
      \captionof{algorithm}{Explicit Leapfrog Step}
      \label{alg:alg2}
      \begin{algorithmic}[1]
        \scriptsize
        \STATE \textbf{Inputs:} $ \mathbf{p}, \bm{\omega}, \tilde{\mathbf{p}}, \tilde{\bm{\omega}}, \epsilon, \varOmega $
        \STATE $\mathbf{p} = \mathbf{p} - \frac{\epsilon}{2} \partial_{\bm{\omega}} H(\bm{\omega},\tilde{\mathbf{p}})$
        \STATE  $\tilde{\bm{\omega}} = \tilde{\bm{\omega}} + \frac{\epsilon}{2} \partial_{\tilde{\mathbf{p}}} H(\bm{\omega},\tilde{\mathbf{p}})$
        \STATE $\tilde{\mathbf{p}} = \tilde{\mathbf{p}} - \frac{\epsilon}{2} \partial_{\tilde{\bm{\omega}}} H(\tilde{\bm{\omega}},\mathbf{p})$
        \STATE  $\bm{\omega} = \bm{\omega} + \frac{\epsilon}{2} \partial_{\mathbf{p}} H(\tilde{\bm{\omega}},\mathbf{p})$
        \STATE $ c = \cos(2\varOmega \epsilon), \quad s = \sin(2\varOmega \epsilon)$
        \STATE  $\bm{\omega} = (\bm{\omega} + \tilde{\bm{\omega}} + c(\bm{\omega} - \tilde{\bm{\omega}}) + s( \mathbf{p} - \tilde{\mathbf{p}}))/2 $
        \STATE  $\mathbf{p} = (\mathbf{p} + \tilde{\mathbf{p}} - s(\bm{\omega} - \tilde{\bm{\omega}}) + c( \mathbf{p} - \tilde{\mathbf{p}}))/2 $
        \STATE  $\tilde{\bm{\omega}} = (\bm{\omega} + \tilde{\bm{\omega}} - c(\bm{\omega} - \tilde{\bm{\omega}}) - s( \mathbf{p} - \tilde{\mathbf{p}}))/2 $
        \STATE  $\tilde{\mathbf{p}} = (\mathbf{p} + \tilde{\mathbf{p}} + s(\bm{\omega} - \tilde{\bm{\omega}}) - c( \mathbf{p} - \tilde{\mathbf{p}}))/2 $
        \STATE $\tilde{\mathbf{p}} = \tilde{\mathbf{p}} - \frac{\epsilon}{2} \partial_{\tilde{\bm{\omega}}} H(\tilde{\bm{\omega}},\mathbf{p})$
        \STATE  $\bm{\omega} = \bm{\omega} + \frac{\epsilon}{2} \partial_{\mathbf{p}} H(\tilde{\bm{\omega}},\mathbf{p})$
        \STATE $\mathbf{p} = \mathbf{p} - \frac{\epsilon}{2} \partial_{\bm{\omega}} H(\bm{\omega},\tilde{\mathbf{p}})$
        \STATE  $\tilde{\bm{\omega}} = \tilde{\bm{\omega}} + \frac{\epsilon}{2} \partial_{\tilde{\mathbf{p}}} H(\bm{\omega},\tilde{\mathbf{p}})$
      \end{algorithmic}
   \end{algorithm}
\end{minipage}


\section{Experiments}\label{sec:exp}

\subsection{Bayesian logistic regression}

The purpose of this illustrative example is to demonstrate both the benefits of RMHMC, and that Explicit RMHMC is significantly faster than Implicit RMHMC. We start with the convex problem of Bayesian logistic regression as the metric tensor is positive semi-definitive and hence well behaved. We define our Hamiltonian as in Equation~\ref{eq:ham}, where the log joint probability is given by
\begin{equation}
\mathcal{L}(\bm{\omega}) = \sum_i^N \{ y_i \log(\bm{\omega}^{\top}\mathbf{x}_i ) + (1-y_i) \log(1-\bm{\omega}^{\top}\mathbf{x}_i )\} + \frac{\beta}{2} \bm{\omega}^{\top} \bm{\omega},
\end{equation}
where the quadratic term in the model parameters corresponds to the prior. We must be careful when defining the precision term $\beta$, as the influence of the prior on the sampling can be significant as we show in Appendix \ref{ap:prior}. Also, we allow the data pairs $\{y, \mathbf{x}\}_{i=1}^N$ to be augmented such that $x_0=1$, which allows $\omega_0$ to correspond to the bias term. We then generate the data by separately sampling $\hat{\bm{\omega}} \sim \mathcal{N}(\mathbf{m}_{\bm{\omega}}, \bm{\Sigma}_{\bm{\omega}} )$ and building a data set through using the logistic $y_i = 1/ \left(1+\exp( -\hat{\bm{\omega}}^{\top}\mathbf{x}_i) \right)$.

\subsubsection{1D example}
\begin{figure}
  \centering
  \includegraphics[width=\textwidth]{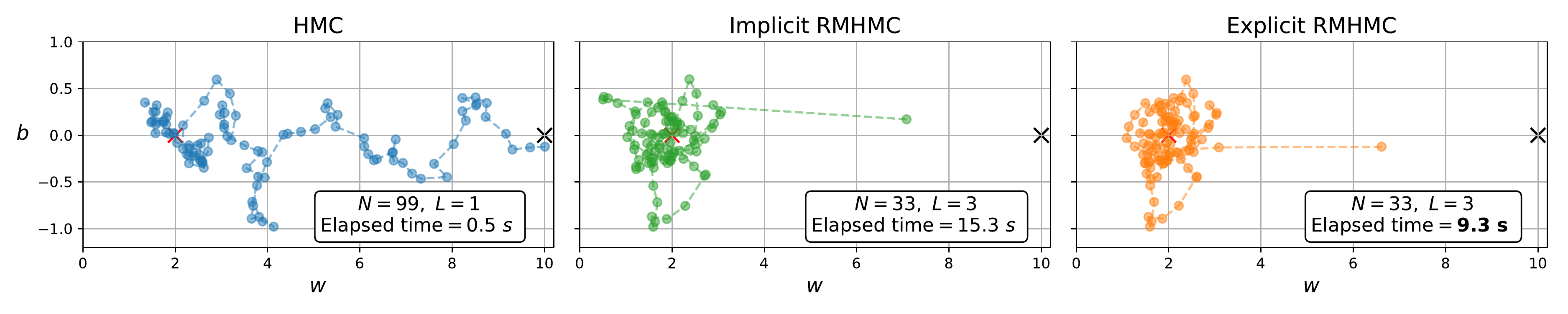}
  \caption{Comparison across the three Hamiltonian-based integration schemes, where all inference engines provide the same number of samples and are initialised with the same weight. This is a logistic regression example, where the weight space is two dimensional and samples are drawn from $\mathcal{N}([2,0],0.3\mathbf{I})$. The red cross marks the known mean from the generative model. This figure shows how the geometric knowledge of the space in both Riemannian models accelerates the particle to the true distribution, whereas the standard HMC model takes a more circuitous route. We highlight the elapsed time, which show a the advantage of Explicit RMHMC.}
  \label{fig:log_reg_1d}
\end{figure}

In a simple example, where $\bm{\omega} \in \mathbb{R}^2$ consists of a weight $w$ and a bias term $b$, we show how RMHMC is able to take advantage of the geometry of the parameter space, whereas HMC does not. We see this in Figure \ref{fig:log_reg_1d}, where we purposely initialise all samplers with $w=10$ and $b=0$ in order to show how each implementation manages when heading towards the known means of $w=2$ and $b=0$. There is a clear difference between the standard HMC sampler and both Riemannian samplers. The incorporation of geometry in both RMHMC schemes avoids the slower route that the HMC scheme follows. Furthermore HMC is renowned for being sensitive to the hyperparameters, such as the step size and the number of leapfrog steps. Therefore despite the fact that RMHMC still requires optimisation of these hyperparameters, it is less sensitive since the step size is automatically adapted according to the curvature of the space.

However, despite the obvious advantages of RMHMC, it comes at a severe computational cost. On the same hardware, the samples in Figure \ref{fig:log_reg_1d} take 0.5 s, 15.3 s and 9.3 s for HMC, Implict RMHMC and Explicit RMHMC respectively.\footnote{CPU: Intel i7-8700k; GPU: NVIDIA TITAN Xp; RAM: 32 GB} This is when allowing the fixed-point routine to take a maximum number of six iterations. Therefore although RMHMC is expensive, even this small example shows that Explicit RMHMC can help in reducing the computational cost.

\subsubsection{Influence of the prior in a 10D example}
Although well known in Bayesian modelling, we use a 10D example to highlight the importance of the prior when designing the sampling scheme in HMC. We find that large values of $\beta$ focus the samples near the zero-mean prior, whereas for a smaller value of $\beta$, we see the samples moving towards $\mathbf{m}_{\bm{\omega}}$. We display these results in Figure \ref{fig:log_reg_prior} in Appendix \ref{ap:prior} in order to highlight how the prior influences results. In addition, when setting the hyperparameters, we initialise all schemes with a long trajectory length and use the autocorrelation to set $L$. We aim to reduce the autocorrelation to result in a larger effective sample size. Appendix \ref{ap:log_reg} provides further details.

\subsubsection{The significance of $\varOmega$}
We cannot introduce a new parameter $\varOmega$ without giving some intuition and empirical example of its influence in practice. We are able to show that as we vary the value of $\varOmega$, small values below a certain threshold display poor long-term performance, whereas above a threshold, we see a long-term performance consistent with the implicit integration scheme. We also see that when $\varOmega$ is set too large, the long-term performance once again degrades. These results are consistent with \citet{pihajoki2015explicit} and \citet{tao2016explicit}, and we refer to Figure \ref{fig:binding} in Appendix \ref{ap:omega}.

\subsection{Inference in a Bayesian hierarchical model}

We now introduce the funnel distribution, which is defined as
$$\prod_i\mathcal{N}(\mathbf{x}_i\vert 0, \exp\{-v\})\mathcal{N}(v\vert 0, 9). $$
This was first introduced by \citet{neal2003slice} and is a good example of a model that is challenging for Bayesian inference. Additionally, the marginal distribution of $v$ is $\mathcal{N}(0,9)$, thus allowing us to make comparisons to the known distribution. 
To demonstrate the advantages of RMHMC over HMC, we use the KL-divergence, $\mathrm{D}_{\mathrm{KL}}(p(v)\Vert q(v))$, as our metric of performance, as well as comparing the wall-clock time. To define $q(v)$, we take the first and second moments of the samples to define our approximation to the known normal distribution $p(v)$. The KL-divergence is appropriate because in a real application we would only have access to $q$, and it is in our interest to know how much information is lost if we rely on $q$ instead of $p$.

We compare four Hamiltonian-based Monte Carlo schemes, where our baseline is the HMC implementation of the No-U-Turn Sampler \citep{hoffman2014no}, which has been set to adapt its acceptance rate to between $0.65 - 0.9$.\footnote{The same parameter settings are followed as in \citet{hoffman2014no}. The desired acceptance rate is set to $0.75$, but this results in large step sizes that cause trajectories to traverse into numerically unstable areas of the log probability space. Therefore the adapted acceptance rate is higher than would normally be desired, although results in the same poor performance.} We further provide an implementation of HMC that has been tuned with knowledge of the true marginal. Importantly, we compare \textit{implicit RMHMC} with our new \textit{explicit RMHMC} sampler. We further highlight that for both Riemannian schemes the Fisher information matrix is no longer guaranteed to be positive semi-definite as it was for Bayesian logistic regression. Therefore we filter out negative eigenvalues by implementing \textit{SoftAbs}, as in \citet{betancourt2013general}, the details of which are available in Appendix \ref{ap:softabs}.

\begin{figure}[h!]
    \centering
    \begin{subfigure}[t]{0.45\textwidth}
        \centering
        \includegraphics[height=1.2in]{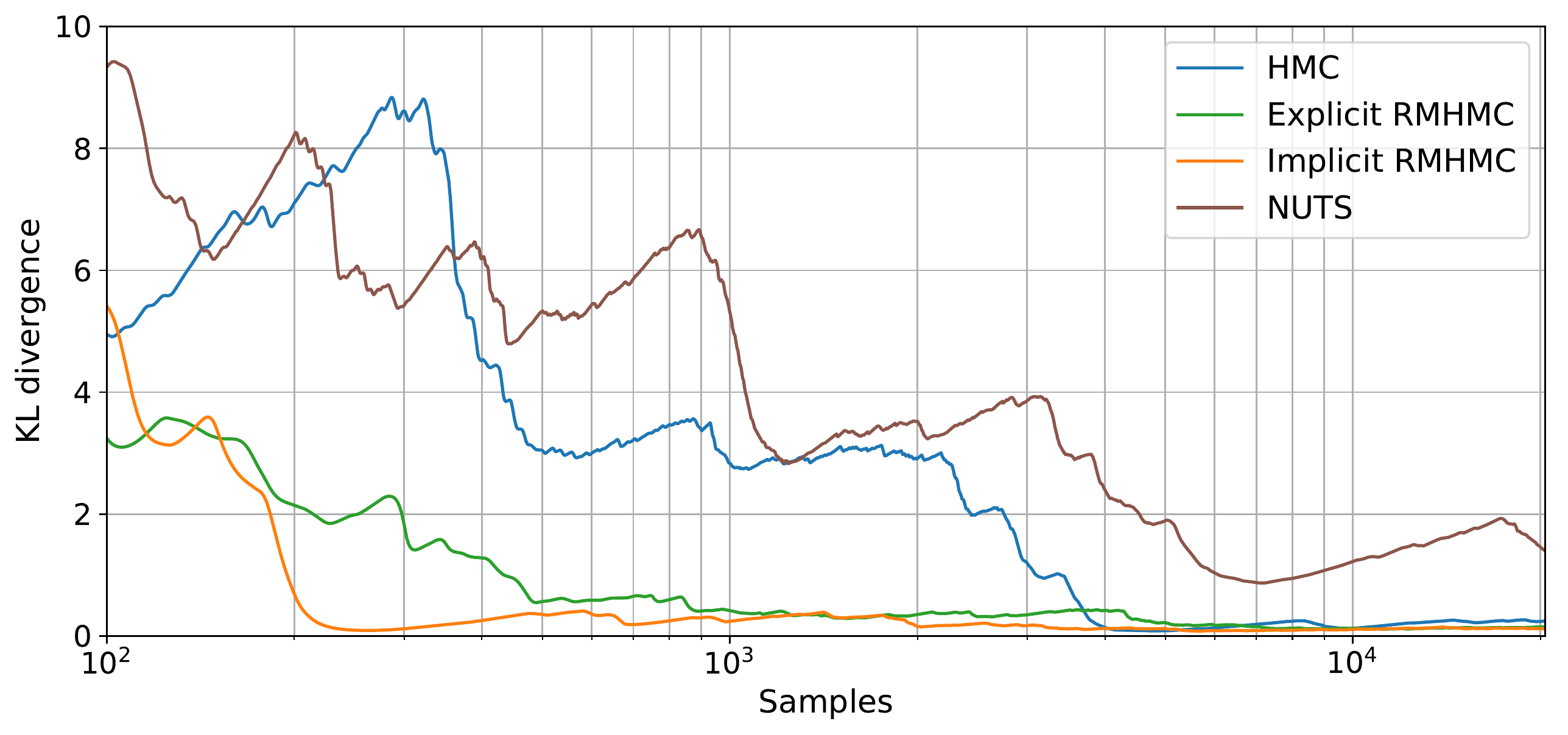}
        \subcaption{KL divergence versus samples.}\label{fig:funnel_kl}
    \end{subfigure}%
    ~ 
    \begin{subfigure}[t]{0.45\textwidth}
        \centering
        \includegraphics[height=1.2in]{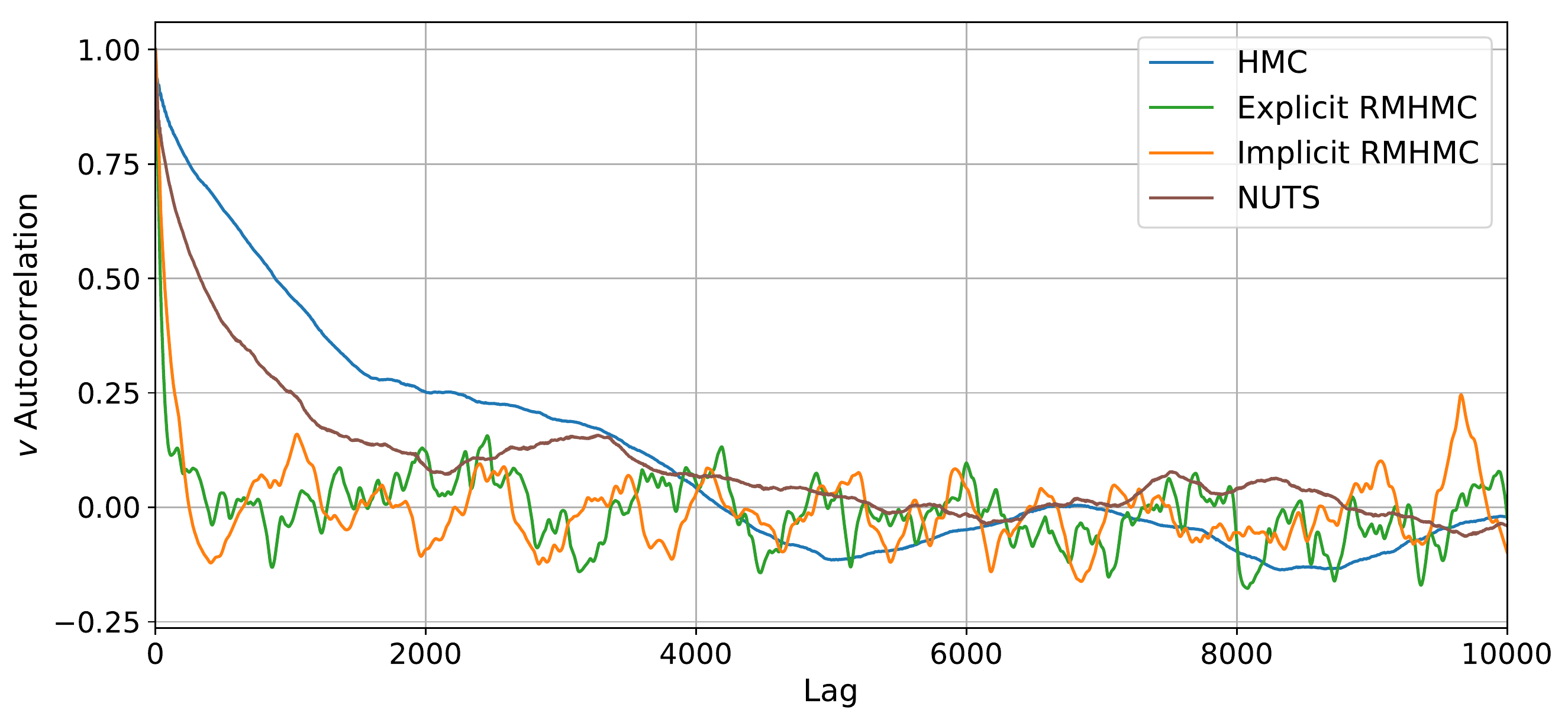}
        \subcaption{Autocorrelation of $v$ versus the lag.}\label{fig:funnel_acf}
    \end{subfigure}
    \caption{Funnel experiment sampler diagnostics.}\label{fig:funnel_diag}
\end{figure}

\begin{table}
\small
  \caption{$10 +1$ dimensional funnel: comparison across integration schemes for inference in the funnel distribution. RMHMC outperforms HMC, even when HMC is optimised with a posteriori information. Furthermore explicit RMHMC achieves comparable performance to implicit RMHMC in almost a third of the time.}
  \label{tab:funnel}
  \centering
  \begin{tabular}{lrrrrrrr}
    \toprule
    Scheme   & $\mathrm{D}_{\mathrm{KL}}$
    & Wall Time & Acc. Rate & $N$ & $L$ & $\epsilon$ & $\alpha$ \\
    \midrule
    HMC - NUTS & $1.109$  & $42$ min  & $0.87$ & $10^5$ & $25$ & $0.3896$ & -   \\
    Implicit RMHMC & $0.130$  &$3$ hr $10$ min & $0.93$ & $10^3$ & $25$ & $0.1500$ & $10^{6}$  \\
    Explicit RMHMC, $\varOmega=10$   & $0.142$  & $1$ hr $12$ min & $0.81$ & $10^3$ & $25$ & $0.1400$ & $10^{6}$  \\
    \midrule
HMC (Hand-Tuned for KL) & $0.270$  & $38$ min & $0.97$ & $10^5$ & $25$ & $0.2$ & -\\
    \bottomrule
  \end{tabular}
  \vspace{-0.1in}
\end{table}

\begin{figure}[h!]
  \centering
  \includegraphics[width=0.9\columnwidth]{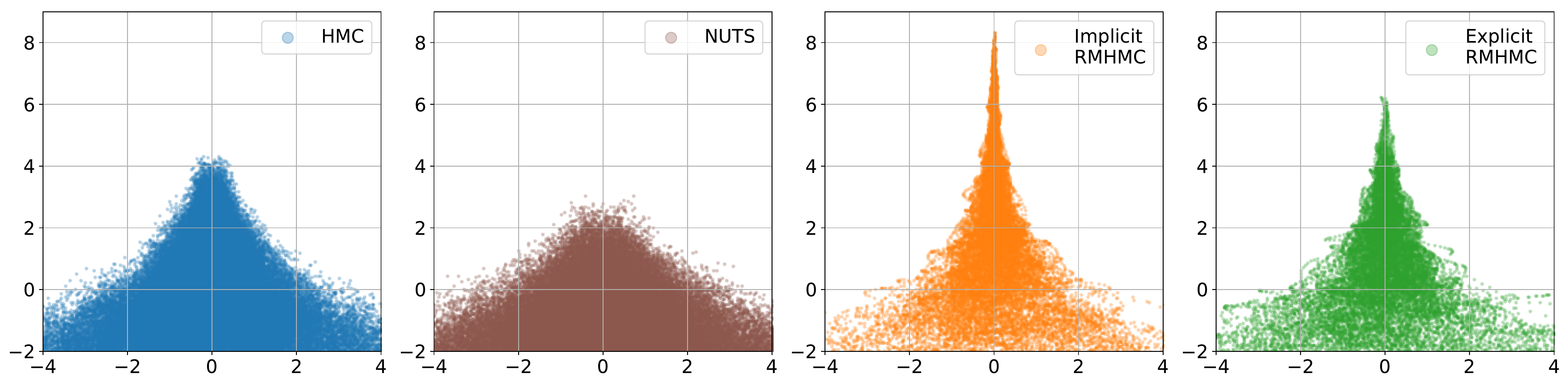}
  \caption{Comparison across Hamiltonian-based schemes for inference in a Bayesian hierarchical model. The Riemannian-based integrators have knowledge of the model's complex geometry and therefore more efficiently explore the space and are able to sample in the narrow part of the funnel. Our explicit integration scheme gives a significant speed-up over the previously used implicit scheme for the same number of samples. The No-U-Turn Sampler (NUTS) adapts its step size to a value that prevents it from traversing up the neck of the funnel.}
  \label{4:fig:funnel_samples}
\end{figure}

\begin{figure}[h!]
  \centering
  \includegraphics[width=0.9\columnwidth]{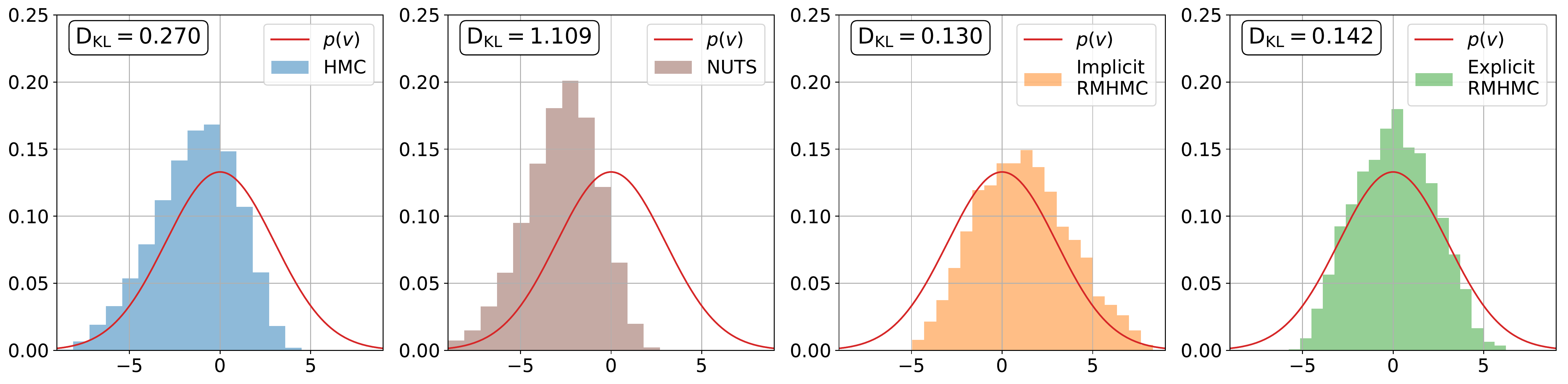}
  \caption{The estimated marginal distributions of $p(v)$ for all four samplers. Compared to the known distribution (the red line), the Riemannian samplers provide samples that appear less biased by the narrowness of the funnel. This leads to a lower (better) KL divergence. }
  \label{4:fig:funnel_hist}
   \vspace{-.3in}
\end{figure}

Table \ref{tab:funnel} shows that both forms of RMHMC outperform HMC in terms of $\mathrm{D}_{\mathrm{KL}}$, despite providing the optimiser for HMC with a posteriori information by hand-tuning with knowledge of the true distribution and by allowing it to take two orders of magnitude more samples. The No-U-Turn Sampler, which adapts for an appropriate acceptance rate leads to too large a step size and prevents it from exploring the narrow neck of the funnel (Figure \ref{4:fig:funnel_samples}). However, the key result is in the wall-clock time, where our new explicit RMHMC achieves comparable $\mathrm{D}_{\mathrm{KL}}(p(v)\Vert q(v))$ to implicit RMHMC in almost a third of the time.\footnote{When the max number of fixed point iterations for Implicit RMHMC was set to $6$ the acceptance rate for the same initial conditions dropped from $0.93$ to $0.50$, therefore the max number was set to $1000$, with the convergence threshold for breaking out of the loop set to $0.001$. At an acceptance rate of $0.50$, the resulting sampler did achieve a similar performance in terms of wall time to Explicit RMHMC but this is clearly not a desirable rate.} We further show both the convergence rate of each sampler in $\mathrm{D}_{\mathrm{KL}}(p(v)\Vert q(v))$ and the autocorrelations in Figure \ref{fig:funnel_diag}, where both RMHMC samplers display comparable performance in convergence and autocorrelation. The faster convergence of $\mathrm{D}_{\mathrm{KL}}(p(v)\Vert q(v))$ for both RMHMC methods demonstrates how the samplers are more efficient at building an empirical representation of $p(v)$, when compared to the other HMC samplers. This is also highlighted by how the autocorrelation between samples plateaus around zero faster for Explicit and Implicit RMHMC. This suggests that the effective sample size of both these samplers is higher than for NUTS and hand-tuned HMC. Finally, the estimated marginal distributions of $p(v)$ are shown in Figure \ref{4:fig:funnel_hist}, where the better performance of the two Riemannian samplers can be seen from the lower KL divergence.

\section{Conclusion}
\label{sec:conclusion}

By introducing an explicit symplectic integration scheme, we have reduced the number of higher-order derivative calculations required per leapfrog step for RMHMC. This results in an almost two times speed up over the previously used implicit integration scheme for RMHMC, with comparable performance.
We have demonstrated these results both for Bayesian logistic regression and a Bayesian hierarchical model, where we have also provided insights as to how one might use explicit RMHMC in practice.

Finally, by reducing the computational burden of RMHMC, we hope to expand the range of models that may currently be too computationally expensive for RMHMC and enable others to do the same with our open-source Python package \texttt{hamiltorch}. We will explore these models in future work, with the objective of providing the tools for improving inference and achieving better uncertainty estimates.

\begin{mybox}
\paragraph{\textbf{Hamiltorch: a PyTorch Python package for sampling}.}
All the Monte Carlo methods implemented in this paper were performed using \texttt{hamiltorch}. This is a Python package we developed that uses Hamiltonian Monte Carlo (HMC) to sample from probability distributions. \texttt{hamiltorch} is built with a PyTorch \citep{paszke2017automatic} backend to take advantage of the innate automatic differentiation.
PyTorch is a machine learning framework that makes modelling with neural networks in Python easier. Therefore since \texttt{hamiltorch} is based on PyTorch, we ensured that \texttt{hamiltorch} was built to sample directly from neural network models (i.e. objects inheriting from the \texttt{torch.nn.Module}).
\\

To use \texttt{hamiltorch} to sample a Bayesian neural network one only needs to define a data set $\{\mathbf{X}, \mathbf{Y}\}$ and a neural network model. PyTorch allows us to run much of the code on a GPU, which results in a large speed increase over running on a CPU. For example, if we use the Pytorch MNIST CNN example \citep{pytorchLenet} based on the LeNet architecture \citep{lecun1998gradient}, for the GPU we are able to achieve $116.33$ samples per second, whereas on the CPU the sampling rate is $13.92$ samples per second. This is for a network that consists of two convolutional layers and two fully connected layers, which contains $431{,}080$ weights (including biases).
\\

Finally, it is simple to switch between integration schemes by changing the argument passed to the sampler. 
Therefore easily switching between HMC and RMHMC is part of our framework. The structure of \texttt{hamiltorch} aims to make performing HMC more scalable and accessible to practitioners, as well as enabling researchers to add their own metrics for RMHMC. For the code base please refer to \url{https://github.com/AdamCobb/hamiltorch} and for a tutorial please refer to \url{https://adamcobb.github.io/journal/hamiltorch.html}.
\end{mybox}

\small

\bibliographystyle{unsrtnat}
\bibliography{paper}


\newpage
\section*{Introducing an Explicit Symplectic Integration Scheme for Riemannian Manifold Hamiltonian Monte Carlo\\(Supplementary Material)}

\appendix
\section{Derivation of distance metric}\label{ap:metric}
A common metric for measuring a distance between distributions is the Kullback--Leibler (KL)divergence,
\begin{equation}\label{eq:KL}
\mathrm{D}_{\mathrm{KL}}\left(p(\mathbf{x}, \bm{\omega}) \vert\vert p(\mathbf{x}, \bm{\omega}')\right) = \iint p(\mathbf{x}, \bm{\omega}) \left(\log p(\mathbf{x}, \bm{\omega}) - \log p(\mathbf{x}, \bm{\omega}')\right) \mathrm{d}\mathbf{x}\mathrm{d}\bm{\omega}.
\end{equation}
If we expand the $\log p(\mathbf{x}, \bm{\omega}')$ term within the integrand using a Taylor expansion around $\bm{\omega}$ such that $\delta\bm{\omega} = \bm{\omega}' - \bm{\omega}$, we obtain
\begin{align}\label{eq:te}
\mathrm{D}_{\mathrm{KL}}\big( p(\mathbf{x}, \bm{\omega}) \vert\vert p(\mathbf{x},  \bm{\omega} + \delta\bm{\omega})\big) &\approx 
\iint p(\mathbf{x}, \bm{\omega}) \log p(\mathbf{x}, \bm{\omega})\mathrm{d}\mathbf{x}\mathrm{d}\bm{\omega} \notag \\
& - \iint  p(\mathbf{x}, \bm{\omega}) \Big( \log p(\mathbf{x}, \bm{\omega}) + \left(\frac{\nabla_{\bm{\omega}} p(\mathbf{x}, \bm{\omega})}{p(\mathbf{x}, \bm{\omega})}\right)^{\top} \delta\bm{\omega} \notag \\ &\quad \quad \quad \quad \quad+ \frac{1}{2} \delta\bm{\omega}^{\top} \nabla_{\bm{\omega}}^2 \log p(\mathbf{x}, \bm{\omega})\delta\bm{\omega}+  \dots \Big) \mathrm{d}\mathbf{x}\mathrm{d}\bm{\omega}.
\end{align}
We note that the first two terms in Equation \eqref{eq:te} cancel and the first order gradient term reduces to zero as follows,
\begin{align}
- \iint  p(\mathbf{x}, \bm{\omega}) \left( \frac{\nabla_{\bm{\omega}} p(\mathbf{x}, \bm{\omega})}{p(\mathbf{x}, \bm{\omega})}\right)^{\top} \delta\bm{\omega}\mathrm{d}\mathbf{x}\mathrm{d}\bm{\omega} = - \iint \nabla_{\bm{\omega}} p(\mathbf{x}, \bm{\omega})^{\top} \delta\bm{\omega}\mathrm{d}\mathbf{x}\mathrm{d}\bm{\omega} =& - \int \nabla_{\bm{\omega}} p(\bm{\omega})^{\top} \delta\bm{\omega}\mathrm{d}\bm{\omega}\notag\\
=&- \nabla_{\bm{\omega}} \mathbf{1}^{\top} \delta \bm{\omega}\notag\\
=&\ 0.
\end{align}
Thus,
\begin{align}
\mathrm{D}_{\mathrm{KL}}\big(p(\mathbf{x}, \bm{\omega}) \vert\vert p(\mathbf{x},  \bm{\omega} + \delta\bm{\omega})\big) \approx & - \iint p(\mathbf{x}, \bm{\omega}) \left(\frac{1}{2} \delta\bm{\omega}^{\top} \nabla_{\bm{\omega}}^2 \log p(\mathbf{x}, \bm{\omega})\delta\bm{\omega}\right)\mathrm{d}\mathbf{x}\mathrm{d}\bm{\omega} \notag\\
= & - \iint p(\mathbf{x}, \bm{\omega}) \left(\frac{1}{2} \delta\bm{\omega}^{\top} \mathbf{G}(\mathbf{x}, \bm{\omega})\delta\bm{\omega}\right)\mathrm{d}\mathbf{x}\mathrm{d}\bm{\omega} \notag , \\
\end{align}
which results in the outcome of Equation \eqref{eq:metric}.

\section{Checking the symplectomorphism of the augmented Hamiltonian binding term}\label{ap:check}
In order for a transformation matrix to be symplectic, it must satisfy the condition \citet{channell1990symplectic}:
\begin{equation}
\mathbf{H}^{\top}\mathbf{J}\mathbf{H} = \mathbf{J}
\end{equation}
where,
\begin{equation}\label{eq:sym_cond}
\mathbf{J} = 
 \begin{bmatrix}
    \mathbf{0} & \mathbf{I}\\
    -\mathbf{I} & \mathbf{0}
\end{bmatrix}
\end{equation}
is a non-singular, skew-symmetric matrix and $\mathbf{H} \in \mathbb{R}^{2D\times 2D}$ is a symplectic matrix. This definition \citep[Chapter~6, Page~183]{hairer2006geometric} is derived from the area preserving properties of the linear mapping $\mathbf{H}$, where the structure of $\mathbf{J}$ originates from the form of the determinant in the transformed space. In nonlinear systems, $\mathbf{H}$ can be the Jacobian of a differentiable function \citep{poincare1899methodes} as shown in \cite[Chapter~6, Page~184]{hairer2006geometric}.

To show how the binding term in Equation \eqref{eq:ham_tao} meets this condition, we can transform the Hamiltonian $\tilde{H}(\bm{\omega},\mathbf{p},\tilde{\bm{\omega}},\tilde{\mathbf{p}})$ to $\hat{H}(\hat{\bm{Q}},\hat{\bm{P}},\bm{\alpha},\bm{\beta})$, where $\hat{\bm{Q}} = \bm{\omega} + \tilde{\bm{\omega}}$, $\hat{\bm{P}} = \mathbf{p} + \tilde{\mathbf{p}}$, $\bm{\alpha} = \bm{\omega} - \tilde{\bm{\omega}}$ and $\bm{\beta} = \mathbf{p} - \tilde{\mathbf{p}}$, giving 
\begin{equation}
\hat{H}(\hat{\bm{Q}},\hat{\bm{P}},\bm{\alpha},\bm{\beta}) =  H_1(\hat{\bm{Q}} +\bm{\alpha}/2 ,\hat{\bm{P}} - \bm{\alpha}/2) + H_2(\hat{\bm{Q}} -\bm{\alpha}/2 ,\hat{\bm{P}} + \bm{\alpha}/2) + \varOmega \frac{1}{2} (\Vert\bm{\alpha}\Vert^2 + \Vert\bm{\beta}\Vert^2 ) + \mathcal{R}
\end{equation}
as shown in \cite{tao2016explicit}, where $\mathcal{R}$ is the perturbative higher order remainder.

Therefore, if we employ this transformation of variables to the binding term $\phi^{\delta}_{\varOmega h}$, the linear system in Equation~ \eqref{eq:tao_integ} corresponding to $\phi^{\delta}_{\varOmega h}$ becomes:
\begin{equation}\label{eq:trans_LS}
\begin{bmatrix}
    \hat{\bm{Q}}\\
    \hat{\bm{P}}\\
    \bm{\alpha}\\
    \bm{\beta}
\end{bmatrix}
=
\begin{bmatrix}
    1 & 0 & 0 & 0\\
    0 & 1 & 0 & 0\\
    0 & 0 & \cos (2\varOmega\delta) & \sin (2\varOmega\delta)\\
    0 & 0 & - \sin (2\varOmega\delta) & \cos (2\varOmega\delta)
\end{bmatrix}
\begin{bmatrix}
    \hat{\bm{Q}}\\
    \hat{\bm{P}}\\
    \bm{\alpha}\\
    \bm{\beta}
\end{bmatrix}
\end{equation}
after the manipulation of the variables from the linear system in \eqref{eq:tao_linear_eq}.

We will now write this as
\begin{equation}
\mathbf{y}' = \mathbf{H} \mathbf{y},
\end{equation}
where $\mathbf{H}$ is the transformation matrix in Equation \eqref{eq:trans_LS}. In order to check this transformation satisfies the condition in Equation \eqref{eq:sym_cond}, we do $\mathbf{H}^{\top}\mathbf{J}\mathbf{H}$. After applying these matrix multiplications and using the trigonometric identity $\sin^2 (x) + \cos^2 (x) = 1$ we see that $\mathbf{H}^{\top}\mathbf{J}\mathbf{H} = \mathbf{J}$ thus confirming $\phi^{\delta}_{\varOmega h}$ is a symplectomorphism.

\section{Hessians with negative eigenvalues}\label{ap:softabs}

In Bayesian logistic regression, we can calculate the Fisher information matrix with knowledge that the metric will be positive semi-definite \citet{girolami2011riemann}. However, once models become more complex, the positive semi-definiteness of the Fisher information metric is no longer guaranteed.

In order to ensure $\mathbf{G}(\bm{\omega})$ is positive semi-definite, we follow the the same \textit{SoftAbs} metric that \cite{betancourt2013general} introduced for RMHMC. This allows us to filter the eigenspectrum by passing all the eigenvalues, $\lambda_d$, through the function
\begin{equation}\label{eq:softabs}
 \tilde{\lambda}_d  = \lambda_d \coth (\alpha \lambda_d).
\end{equation}
This approximates the absolute values of the eigenvalues and therefore ensures that the new metric $\tilde{\mathbf{G}}(\bm{\omega}) = \mathbf{Q} \tilde{\bm{\lambda}} \mathbf{Q}^{\top}$ is positive semi-definite, by introducing a parameter $\alpha$ which controls the smoothness. As $\alpha\rightarrow\infty$ the SoftAbs function becomes the absolute value. In our work we set $\alpha = 10^6$ as in \cite{betancourt2013general}. 

\section{Experiments}\label{ap:exp}

\subsection{Bayesian logistic regression}\label{ap:log_reg}
\begin{figure}[h]
\centering
\begin{subfigure}{.5\textwidth}
  \centering
  \includegraphics[width=0.93\textwidth]{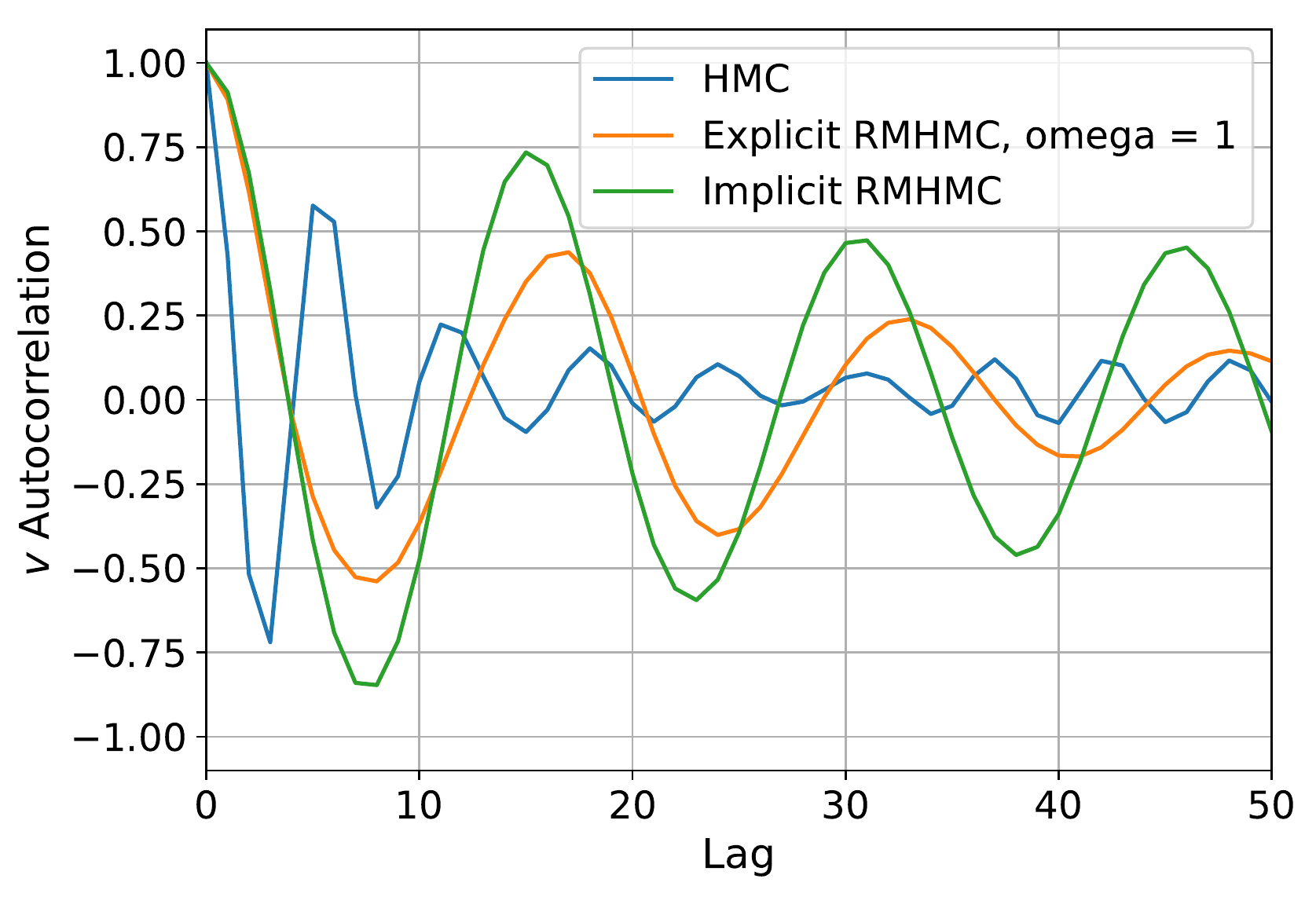}
    \caption{$L = 30$ and results in oscillatory behaviour.}\label{fig:acf_log_reg}
\end{subfigure}%
\begin{subfigure}{.5\textwidth}
  \centering
  \includegraphics[width=0.93\textwidth]{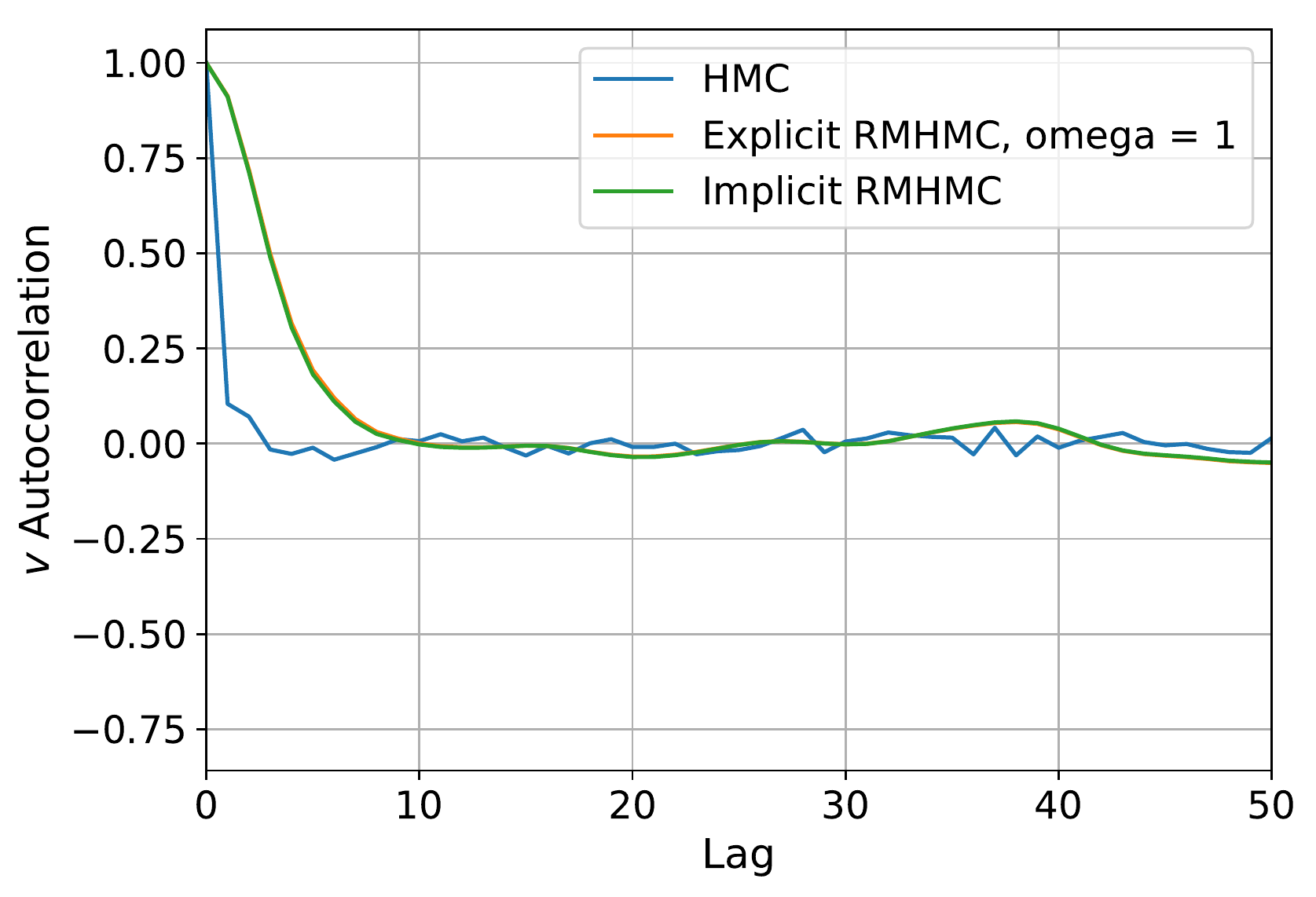}
    \caption{$L$ is reduced to less than half the period for each respective scheme.}\label{fig:acf_log_reg_tuned}
\end{subfigure}
\caption{Autocorrelation for Bayesian logistic regression, where $\bm{\omega} \in \mathbb{R}^{11}$.}
\label{fig:test}
\end{figure}

To tune the trajectory length, $L$ we started with $L=30$ and ran for $100$ iterations. We then looked at the autocorrelation in Figure \ref{fig:acf_log_reg} and set $L$ to less than half the period of the oscillations for each respective inference scheme (HMC: $L=1, \epsilon=0.2$. Implicit RMHMC: $L=3, \epsilon=0.04$. Explicit RMHMC: $L=3, \epsilon=0.04$.). This led us to Figure \ref{fig:acf_log_reg_tuned}, where the autocorrelation is significantly reduced.

\subsubsection{Dependence on the prior and the initialisation of RMHMC}\label{ap:prior}

\begin{figure}
     \centering
     \begin{subfigure}[b]{0.7\textwidth}
         \centering
         \includegraphics[width=\textwidth]{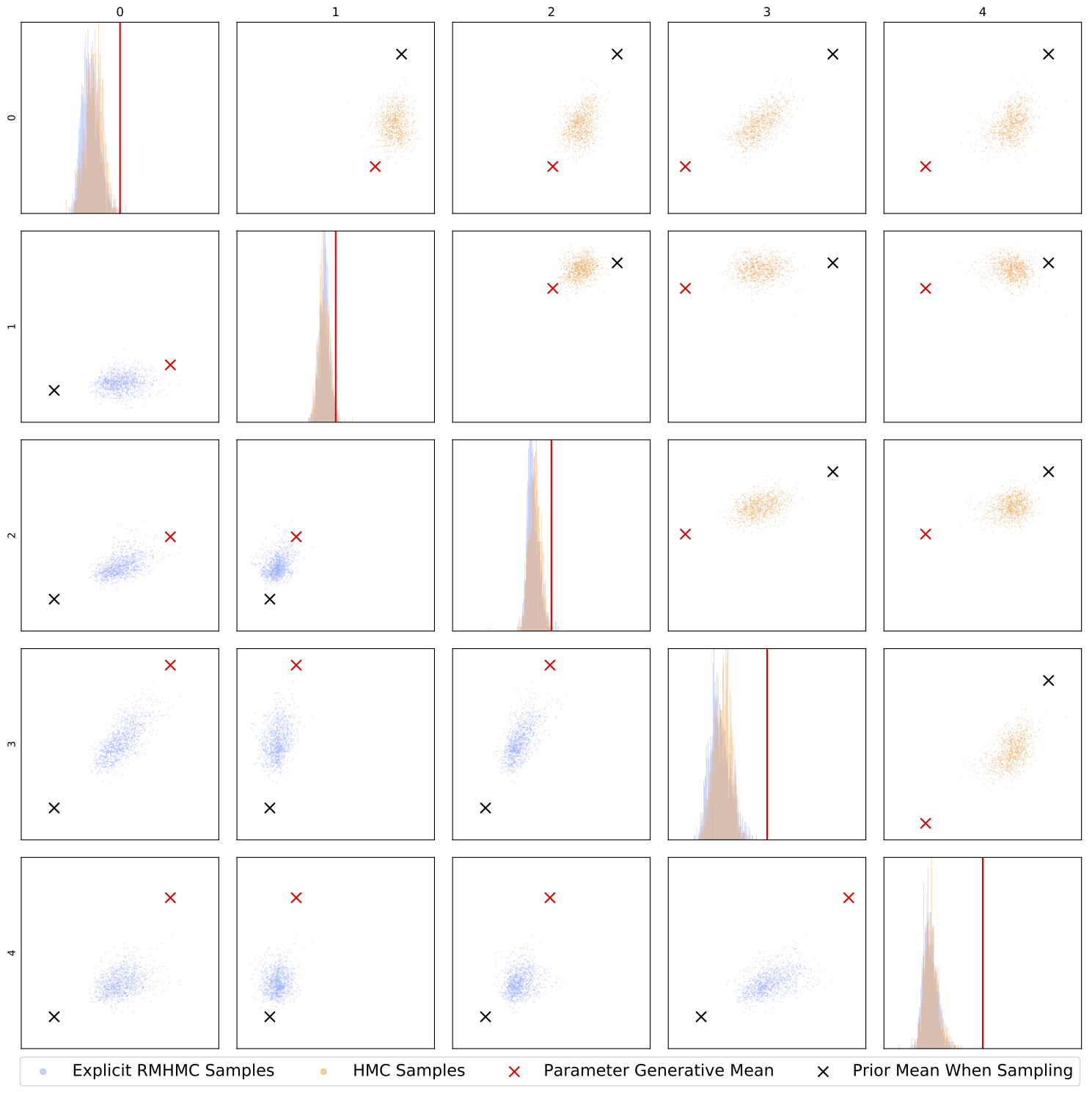}
         \caption{Weak Prior, $\beta = 0.2$.}
         \label{fig:weak_prior}
     \end{subfigure}
     \begin{subfigure}[b]{0.7\textwidth}
         \centering
         \includegraphics[width=\textwidth]{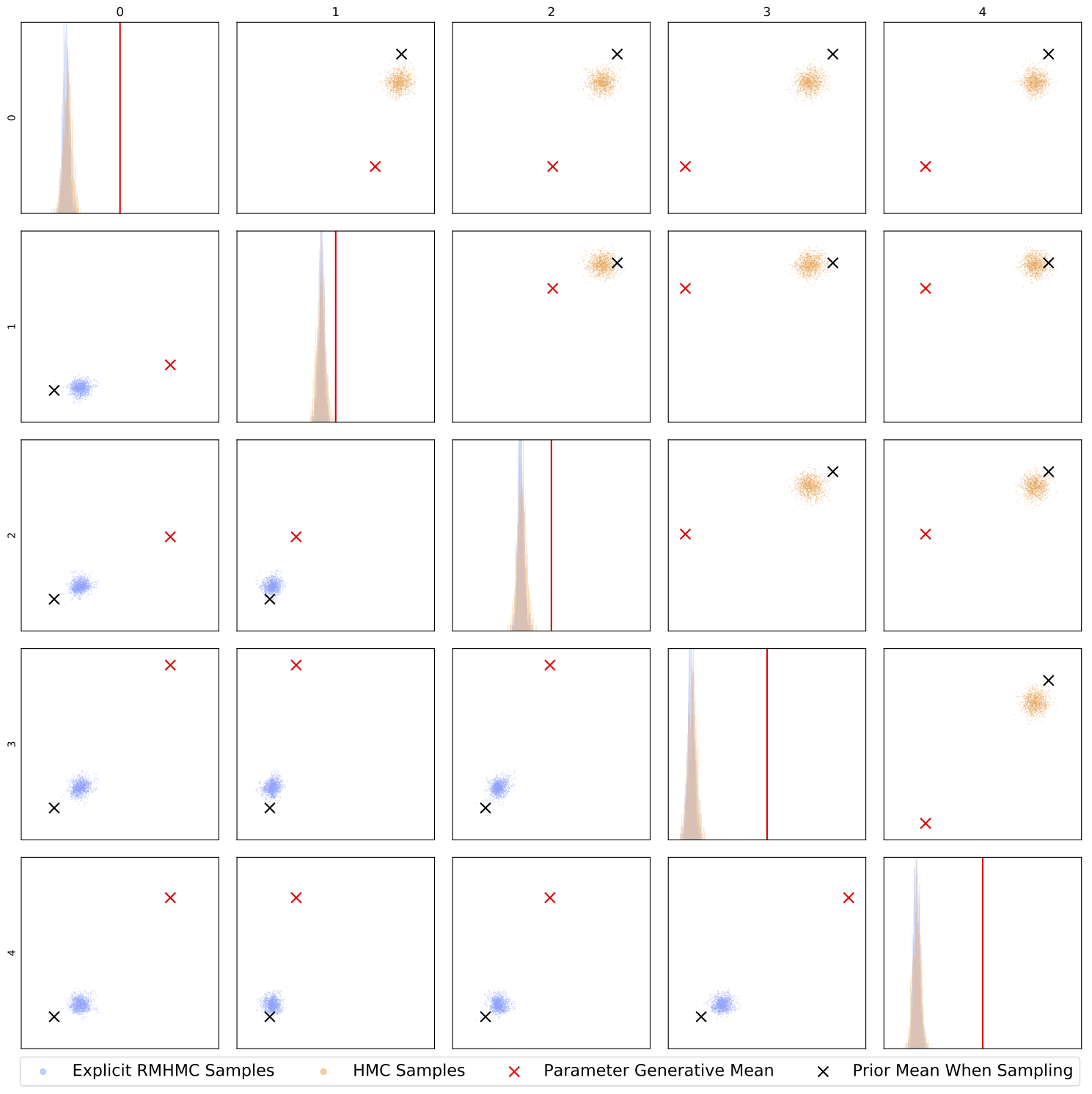}
         \caption{Strong Prior, $\beta = 10$.}
         \label{fig:strong_prior}
     \end{subfigure}
        \caption{A comparison between Explicit RMHMC and HMC for Bayesian logistic regression, where $\bm{\omega} \in \mathbb{R}^{11}$. The purpose of these plots is to both demonstrate how HMC and RMHMC compare in higher dimensions and to show how the prior affects the sampling. We note that we display the results of Explicit RMHMC rather than Implicit RMHMC due to their similar performance. The HMC samples tend to be more spread than the RMHMC samples, however the RMHMC samples lead to slightly better accuracy performance on the logistic regression task.}
        \label{fig:log_reg_prior}
\end{figure}

In this logistic regression example the assumed prior over the combined parameter vector, $\bm{\omega}$, is given by $\mathcal{N}(\mathbf{0},\beta \mathbf{I})$. We note that this is compared to the known generative model $\hat{\bm{\omega}} \sim \mathcal{N}(\mathbf{m}_{\bm{\omega}}, \bm{\Sigma}_{\bm{\omega}} )$ when we build the data set. The value of $\beta$ defines the confidence in the prior. We expect that for large values of $\beta$, the structure of the prior will dominate the when inferring the parameters, whereas small values of $\beta$ are expected to result in broader, less-informative priors that reduce in their influence in the presence of more data. We display this behaviour in Figure \ref{fig:log_reg_prior}, where we display the samples for five of the eleven parameters.

Furthermore, it often benefits the RMHMC schemes to be initialised by a cheap HMC burn-in to ensure that the linear algebra associated with the Hessian is numerically stable (we found that far away initialisations caused numerical issues due to the Hessian becoming singular). 

\subsubsection{Implications of the binding term on performance}\label{ap:omega}

\begin{figure}[h]
  \centering
  \includegraphics[width=0.7\columnwidth]{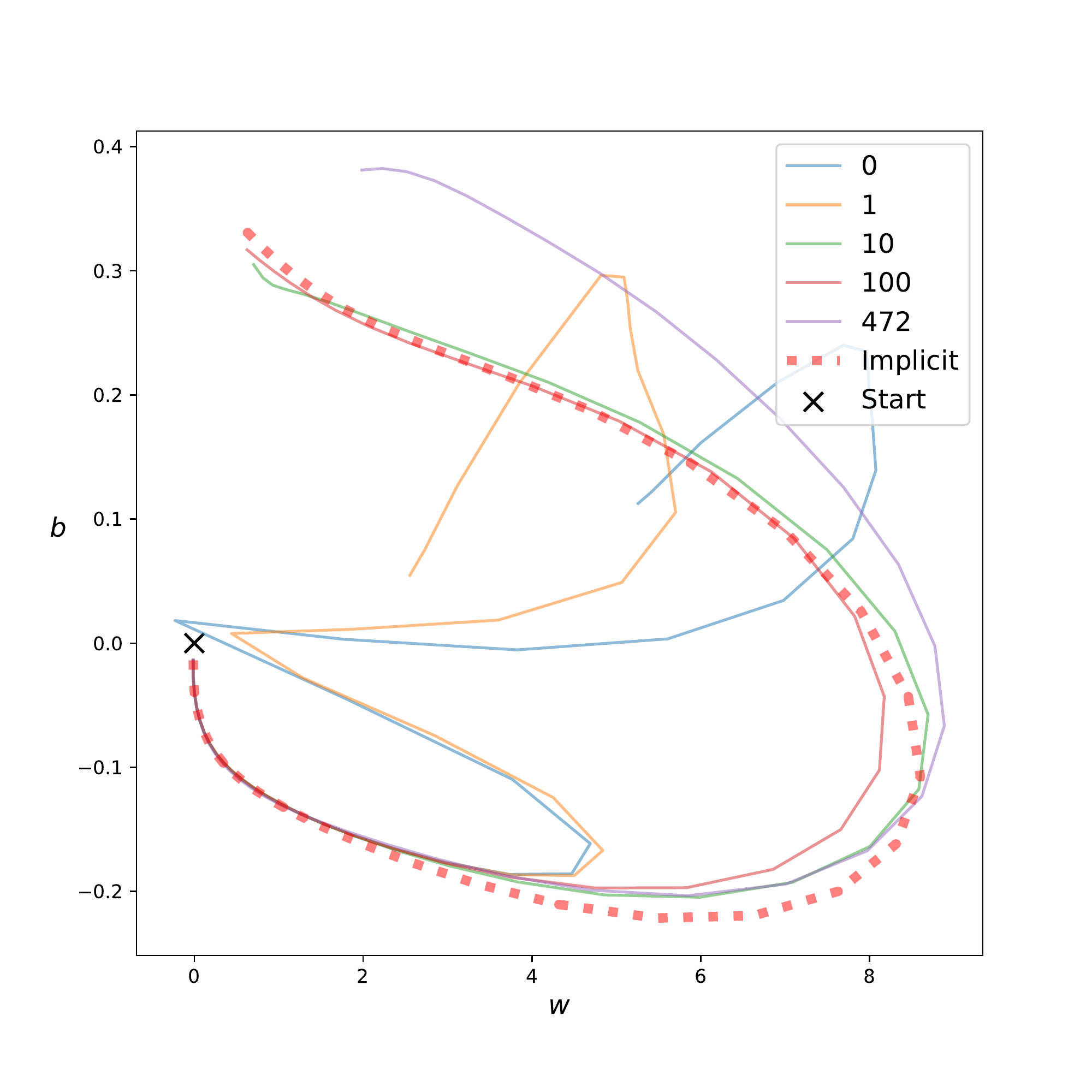}
  \caption{Long-term performance of explicit RMHMC when comparing to Implicit RMHMC for a single trajectory ($L=40, \epsilon= 0.012$). Small values of $\varOmega$, as indicated in the legend, diverge shortly after the initial conditions, as well as the largest value.}
  \label{fig:binding}
\end{figure}

For the single 1D case, such that $w \in \mathbb{R}^{1}$ and $b \in \mathbb{R}^1$, we can test how the performance of the binding term of the explicit integrator ($\varOmega$) affects long term performance.
If we initialise the samplers at the location $[0,0]$ in parameter space and set the trajectory length to $40$, we can observe how different values of $\varOmega$ affect the trajectory. In Figure \ref{fig:binding}, we compare the implicit integration scheme in Equation \eqref{eq:leapfrog} to various explicit schemes with different values for $\varOmega$. We treat the dashed red line, Implicit RMHMC, as the ground truth and show how all other explicit RMHMC trajectories compare. It is clear from this simple example, that $\varOmega$ has an important implication on the long term performance of RMHMC. This was also pointed out in the original paper by \citet{tao2016explicit}, whereby $\varOmega$ must be larger than some threshold $\varOmega_0$ but also small enough such that the error bound $\mathcal{O}(N\delta^{2}\varOmega)$ remains small. Figure \ref{fig:binding} shows what this means in practice.

In fact the purpose of the additional binding term was to improve on the long term performance of the integration scheme introduced by \cite{pihajoki2015explicit}, which is equivalent to $\varOmega=0$. We can clearly see this is the case, as the trajectory corresponding to $\varOmega=0$ in Figure \ref{fig:binding} diverges from the implicit scheme shortly after traversing away from the initial condition. Furthermore, the largest value we were able to set for the binding term before the trajectory was rejected, $\varOmega=472$, also diverges from the implicit scheme. 

\end{document}